\documentclass[10pt,twocolumn,letterpaper]{article}
 
\usepackage{cvpr}
\usepackage{times}
\usepackage{epsfig}
\usepackage{graphicx}
\usepackage{makecell}
\usepackage{amsmath}
\usepackage{amssymb}
\usepackage{multirow}
\usepackage{caption} %
\usepackage{paralist} %
\usepackage{booktabs} %

\long\def\ignorethis#1{}

\renewcommand{\etal}{et~al\mbox{.}}

\newcommand{\heading}[1]{\vspace{0mm}\noindent\textbf{#1}}

\newcommand{\figref}[1]{Figure~\ref{fig:#1}}%
\newcommand{\tabref}[1]{Table~\ref{tab:#1}} %
\newcommand{\eqnref}[1]{Equation~\eqref{eq:#1}}
\newcommand{\secref}[1]{Section~\ref{sec:#1}}

\newcommand{\tb}[1]{\textbf{#1}}

\renewcommand{\paragraph}[1]{\vspace{1mm} \noindent\textbf{#1}}

\newcommand{\figmargin}{\vspace{-2mm}}
\newcommand{\figcapmargin}{\vspace{0mm}}

\newcommand{\ignore}[1]{}   %

\newcommand{\CR}[1]{{#1}}

\newcommand{\mpage}[2]
{
\begin{minipage}{#1\linewidth}\centering
#2
\end{minipage}
}

\usepackage{array}
\newcolumntype{L}[1]{>{\raggedright\let\newline\\\arraybackslash\hspace{0pt}}m{#1}}
\newcolumntype{C}[1]{>{\centering\let\newline\\\arraybackslash\hspace{0pt}}m{#1}}
\newcolumntype{R}[1]{>{\raggedleft\let\newline\\\arraybackslash\hspace{0pt}}m{#1}}

\usepackage{algorithm}
\usepackage[noend]{algpseudocode}

\usepackage[pagebackref=true,breaklinks=true,letterpaper=true,colorlinks,linkcolor=blue,citecolor=blue,urlcolor=blue,bookmarks=false]{hyperref}

\cvprfinalcopy %

\def\cvprPaperID{2197} %

\ifcvprfinal\fi
\begin{document}

\title{Learning to See Through Obstructions}

\author{
Yu-Lun Liu$^{1,4}$\quad
Wei-Sheng Lai$^{2}$\quad
Ming-Hsuan Yang$^{2,3}$\quad
Yung-Yu Chuang$^{1}$\quad
Jia-Bin Huang$^{5}$
\\
$^{1}$National Taiwan University \quad
$^{2}$Google \quad
$^{3}$UC Merced \quad
$^{4}$MediaTek Inc.  \quad
$^{5}$Virginia Tech \\
{\small\url{https://www.cmlab.csie.ntu.edu.tw/~yulunliu/ObstructionRemoval}}
}

\twocolumn[{%
\renewcommand\twocolumn[1][]{#1}%
\vspace{-4mm}
\maketitle
\vspace{-5mm}
\begin{center}
    \captionsetup{type=figure}
    \footnotesize
    \begin{minipage}[c]{1\textwidth}
    \centering
        \fbox{\includegraphics[width=0.32\textwidth]{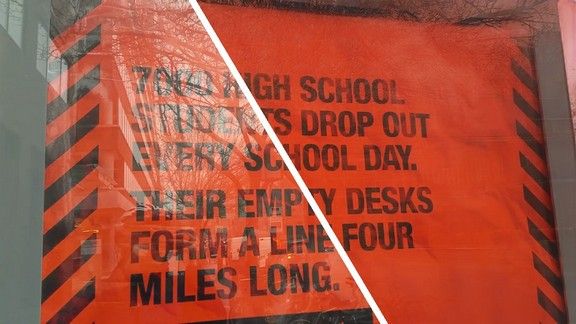}} \hfill
        \fbox{\includegraphics[width=0.32\textwidth]{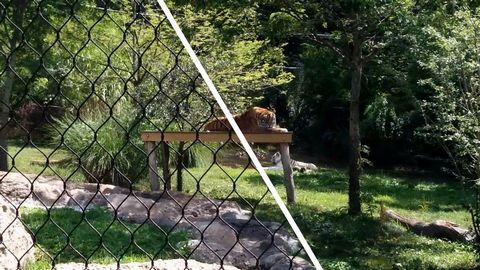}} \hfill
        \fbox{\includegraphics[width=0.32\textwidth]{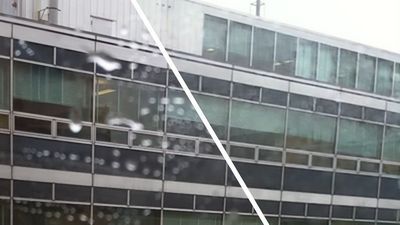}}
    \end{minipage}
    \\
    \figcapmargin
    \mpage{0.32}{(a) Reflection removal} \hfill
    \mpage{0.32}{(b) Fence removal} \hfill
    \mpage{0.32}{(c) Raindrop removal}
    \figmargin
    \caption{
    \tb{Seeing through obstructions.} 
    We present a learning-based method for recovering clean images from a given short sequence of images taken by a moving camera through obstructing elements such as (a) windows, (b) fence, or (c) raindrop.
    }
    \label{fig:teaser}
\end{center}
}]

\begin{abstract}
We present a learning-based approach for removing unwanted obstructions, such as window reflections, fence occlusions or raindrops, from a short sequence of images captured by a moving camera.
Our method leverages the motion differences between the background and the obstructing elements to recover both layers.
Specifically, we alternate between estimating dense optical flow fields of the two layers and reconstructing each layer from the flow-warped images via a deep convolutional neural network. 
The learning-based layer reconstruction allows us to accommodate potential errors in the flow estimation and brittle assumptions such as brightness consistency.
We show that training on synthetically generated data transfers well to real images.
Our results on numerous challenging scenarios of reflection and fence removal demonstrate the effectiveness of the proposed method.
\end{abstract}

\section{Introduction}\label{sec:intro}

Taking clean photographs through reflective surfaces (such as windows) or occluding elements (such as fences) is challenging as the captured images inevitably contain both the scenes of interests and the obstructions caused by reflections or occlusions. 
An effective solution to recover the underlying clean image is thus of great interest for improving the quality of the images captured under such conditions or allowing computers to form a correct physical interpretation of the scene, e.g., enabling a robot to navigate in a scene with windows safely.

Recent efforts have been focused on automatically removing unwanted reflections or occlusions from \emph{a single image}~\cite{arvanitopoulos2017single, fan2017generic, jin2018learning, jonna2016deep, park2010image, wei2019single, yang2018seeing, zhang2018single}.
These methods either leverage the ghosting cues~\cite{shih2015reflection} or adopt learning-based approaches to capture the prior of natural images~\cite{fan2017generic, jin2018learning, wei2019single, yang2018seeing, zhang2018single}. 
While impressive results have been shown, separating the clean background from reflection/occlusions is fundamentally ill-posed and often requires a high-level semantic understanding of the scene to perform well.
In particular, the performance of learning-based methods degrades significantly for out-of-distribution images. 

To tackle these challenges, \emph{multi-frame approaches} have been proposed for reflection/occlusion removal.
The core idea is to exploit the fact that the background scene and the occluding elements are located at different depths with respect to the camera (e.g., virtual depth of window reflections). 
Consequently, taking multiple images from a slightly moving camera reveals the motion differences between the two layers~\cite{be2008blind, gai2009blind, guo2014robust, li2013exploiting, liu2008sift, szeliski2000layer}. 
A number of approaches exploit such cues for reflection or fence removal from a video~\cite{alayrac2019visual, be2008blind, du2018accurate, gai2009blind, guo2014robust, li2013exploiting, liu2008sift, nandoriya2017video, sinha2012image, szeliski2000layer}.
Xue et al.~\cite{xue2015computational} propose a unified computational framework for obstruction removal and show impressive results on several natural sequences.
The formulation, however, requires a computationally expensive optimization process and relies on strict assumptions of brightness constancy or accurate motion estimation.
To alleviate these issues, recent work~\cite{alayrac2019visual} explores model-free methods by using a generic 3D convolutional neural network (CNN).
Yet, the CNN-based methods do not produce results with comparable quality as optimization-based algorithms on real input sequences.

In this work, we propose a multi-frame obstruction removal algorithm that exploits the advantages of both optimization-based and learning-based methods. 
Inspired by the optimization-based approach~\cite{xue2015computational}, the proposed algorithm alternates between the dense motion estimation and the background/obstruction layer reconstruction steps in a coarse-to-fine manner. 
The explicit modeling of dense motion allows us to progressively recover detailed content in the respective layers.
Instead of relying on hand-crafted objectives for solving the layers, we exploit the learning-based method for fusing flow-warped images to accommodate potential violations of brightness constancy and errors in flow estimation.
We train our fusion network using a synthetically generated dataset and demonstrate it transfers well to unseen real-world sequences.
In addition, we present an online optimization process to further improve the visual quality of particular testing sequences.
Finally, we demonstrate that the proposed method performs favorably against existing algorithms on a wide variety of challenging sequences and applications.

\CR{
Our framework builds upon the optimization-based formulation of \cite{nandoriya2017video, xue2015computational} but differs in that our model is purely data-driven and does not rely on classical assumptions such as
brightness constancy~\cite{nandoriya2017video, xue2015computational}, accurate flow fields~\cite{li2013exploiting}, or planar surface~\cite{guo2014robust} in the scene.
When these assumptions are violated (e.g., occlusion/dis-occlusion, motion blur, inaccurate flow), classical approaches may fail to reconstruct clear foreground and background layers.
On the other hand, data-driven approaches learn from diverse training data and can tolerate errors when these assumptions are violated.
}

The contributions of this work include:
\begin{compactitem}
    \item We present a learning-based method that integrates the optimization-based formulation for robustly reconstructing background/obstruction layers. 
    \item We demonstrate that combining model pre-training using synthetically generated data and fine-tuning with real testing sequence (in an unsupervised manner) leads to state-of-the-art performance.
    \item We show our model with minimum design changes can be applied to various obstruction removal problems.
\end{compactitem}

\section{Related work}
\label{sec:related}

\heading{Multi-frame reflection removal.} 
Existing methods often exploit the differences of motion patterns between the background and reflection layers~\cite{guo2014robust,xue2015computational} and impose natural image priors~\cite{gai2011blind,guo2014robust,xue2015computational}.
These methods differ in their way of modeling the motion fields, e.g., SIFT flow~\cite{li2013exploiting}, homography~\cite{guo2014robust}, and dense optical flow~\cite{xue2015computational}.
Recent advances include optimizing temporal coherence~\cite{nandoriya2017video} and learning-based layer decomposition~\cite{alayrac2019visual}.
Compared to learning a generic CNN~\cite{alayrac2019visual}, our method explicitly models the dense flow fields of the background and obstruction layers to obtain sharper and cleaner results on real sequences.

\heading{Single-image reflection removal.} 
A number of approaches have been proposed to remove unwanted reflections with only \emph{one single image} as input.
Existing methods exploit various cues, including ghosting effect~\cite{shih2015reflection}, blurriness caused by depth-of-field~\cite{li2014single,wan2016depth}, image priors (either hand-designed~\cite{arvanitopoulos2017single} or learned from data~\cite{yang2018seeing, zhang2018single}), and the defocus-disparity cues from dual pixel sensors~\cite{punnappurath2019reflection}.
Despite the demonstrated success, reflection removal from a single image remains challenging due to the nature of this highly ill-posed problem and the lack of motion cues.
Our work instead utilizes the motion cues from image sequences captured with a slightly moving camera for separating the background and reflection layers.

\heading{Occlusion and fence removal.}
Occlusion removal aims to eliminate the captured obstructions, e.g., fence or raindrops on an image or sequences, and provide a clear view of the scene.
Existing methods detect fence patterns by exploiting visual parallax~\cite{mu2013video}, dense flow field~\cite{xue2015computational}, disparity maps~\cite{jonna2017stereo}, or using a graph-cut~\cite{yi2016automatic}.
One recent work leverages a CNN for fence segmentation~\cite{du2018accurate} and recovers the occluded pixels using optical flow.
Our method also learns deep CNNs for optical flow estimation and background image reconstruction.
Instead of focusing on fence removal, our formulation is more general and applicable to different obstruction removal tasks.

\heading{Video completion.}
Video completion aims to fill in plausible content in missing regions of a video~\cite{ilan2015survey}, with applications ranging from object removal, full-frame video stabilization, and watermark/transcript removal.
State-of-the-art methods estimate the flow fields in both known and missing regions to constrain the content synthesis~\cite{huang2016temporally,xu2019deep}, and generate temporally coherent results.
The obstruction removal problem resembles a video completion task.
However, the crucial difference is that no manual mask selection is required for removing the fences/obstructions from videos.

\heading{Layer decomposition.}
Image layer decomposition is a long-standing problem in computer vision, e.g., intrinsic image~\cite{bell2014intrinsic,zhou2015learning}, depth, normal estimation~\cite{jeon2014intrinsic},  relighting~\cite{eisemann2004flash}, and inverse rendering~\cite{li2019inverse, sengupta2019neural}.
Our method is inspired by the development of the approaches for these layer decomposition, particularly in the ways of leveraging both the physical image formation constraints and data-driven priors. 

\heading{Online optimization.}
Learning from the test data has been an effective way to reduce the domain discrepancy between the training/testing distributions.
Examples include using geometric constraints~\cite{chen2019self}, self-supervised losses~\cite{sun2019test}, and online template update~\cite{kalal2011tracking}.
Similar to these methods, we apply online optimization to fine-tune our background/obstruction reconstruction network on a particular test sequence to further improve the separation. 
Our unsupervised loss directly measures how well the recovered background/obstruction and the dense flow fields explain all the input frames.

\begin{figure}
    \centering
    \includegraphics[width=1.0\linewidth]{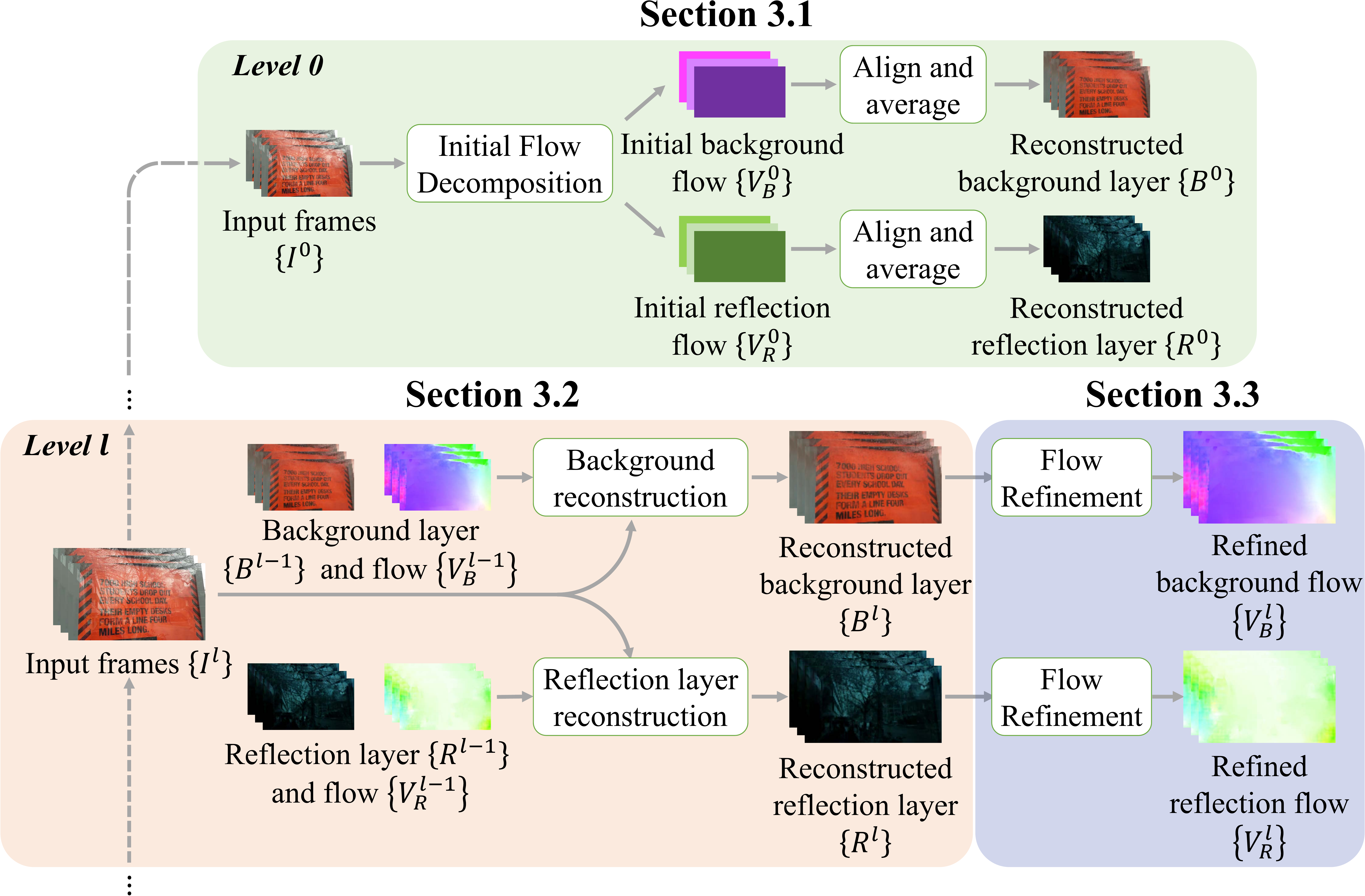}
    \vspace{-3mm}
    \caption{
    \textbf{Algorithmic overview.} 
    \CR{
    We reconstruct the background/reflection layers in a coarse-to-fine manner.
    At the coarsest level, we estimate uniform flow fields for both the background and reflection layers and then reconstruct coarse background/reflection layers by averaging the aligned frames.
    At level $l$, we apply (1) background/reflection layer reconstruction modules to reconstruct the background/reflection layer, and (2) use the PWC-Net to predict the refined flow fields for both layers.
    Our framework progressively reconstructs the background/reflection layers and flow fields until the finest level.}
    }
    \label{fig:archi}
\end{figure}

\section{Proposed Method}
\label{sec:algorithm}
Given a sequence $\{I_t\}^T_{t=1}$ of $T$ frames, the goal is to decompose each frame $I_k$ into two layers, one for the (clean) background and the other for obstruction caused by fense/raindrops/occlusion.
Decomposing an image sequence into background and obstruction layers is difficult as it involves solving two tightly coupled problems: optical flow decomposition and layer reconstruction.
Without a good flow decomposition, the layers cannot be reconstructed faithfully due to the misalignment from inaccurate motion estimation.
On the other hand, without well-reconstructed background and obstruction layers, the optical flow cannot be accurately estimated because of the mixed content.
Due to the nature of this chicken-and-egg problem, there is no ground to start with because we do not have information for both flows and layers.

In this work, we propose to learn deep CNNs to address the challenges.
Our proposed method mainly consists of three modules: 
1) initial flow decomposition, 
2) background and obstruction layer reconstruction, and 
3) optical flow refinement.
Our method takes $T$ frames as input and aims to decompose the keyframe frame $I_k$ into a background layer $B_k$ and reflection layer $R_k$ at a time.
We reconstruct the output images in a coarse-to-fine manner within an $L$-level hierarchy.
First, we estimate the flows at the coarsest level from the initial flow decomposition module (Section~\ref{initial_flow_decomposition}).
We then progressively reconstruct the background/obstruction layers (Section~\ref{background_reflection_layer_reconstruction}) and refine optical flows (Section~\ref{optical_flow_refinement}) until the last level.
\figref{archi} shows an overview of our method.
Our unified framework can be applied to several layer decomposition problems, such as reflection/obstruction/fence/rain removal.
Without loss of generality, we use the reflection removal task as an example to introduce our algorithm. 
We describe the details of the three modules in the following sections.

\subsection{Initial Flow Decomposition}
\label{initial_flow_decomposition}
We first predict the flow for both background and reflection layers at the coarsest level ($l = 0$), which is the essential starting point of our algorithm.
Instead of estimating dense flow fields, we propose to learn a \emph{uniform} motion vector for each layer.
Our initial flow decomposition network consists of two sub-modules: 1) a feature extractor, and 2) a layer flow estimator.
The feature extractor first generates feature maps for all the input frames at a $1/{2^L}\times$ spatial resolution.
Then, we construct a cost volume between frame $j$ and frame $k$ via a correlation layer~\cite{sun2018pwc}:
\begin{equation} \label{eq:cost_volume}
    CV_{jk}(\mathbf{x_1}, \mathbf{x_2}) = c_j(\mathbf{x_1})^\top c_k(\mathbf{x_2}),
\end{equation}
where $c_j$ and $c_k$ are the extracted features of frame $j$ and $k$, respectively, and $\mathbf{x}$ indicates the pixel index.  
Since the spatial resolution is quite small at this level, we set the search range of the correlation layer to only 4 pixels.
The cost volume $CV$ is then concatenated with the feature $c_j$ and fed into the layer flow estimator.

The layer flow estimator uses the global average pooling and fully-connected layers to generate two global motion vectors.
Finally, we tile the global motion vectors into two uniform flow fields (at a $1/{2^L}\times$ spatial resolution):
$\{V^0_{B, j\rightarrow k}\}$ for the background layer
and 
$\{V^0_{R, j\rightarrow k}\}$ for the reflection layer.
We provide the detailed architecture of our initial flow decomposition module in the supplementary material.

\begin{figure*}
    \centering
    \includegraphics[width=0.9\linewidth]{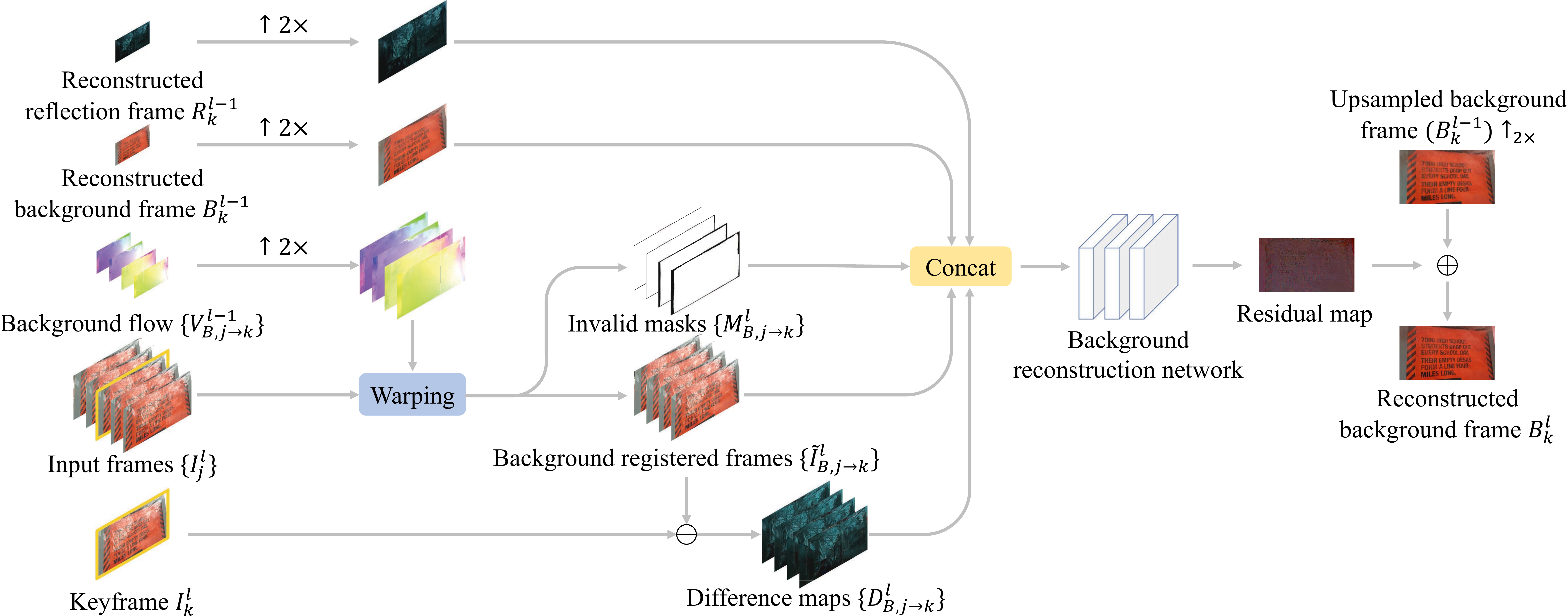}
    \figmargin
    \caption{
        \textbf{Overview of layer reconstruction module.}
        \CR{
        At level $l$, we first upsample the background flows $\{V^{l-1}_{B, j\rightarrow k}\}$ from level $l - 1$ to warp and align the input frames $\{I^l_t\}$ with the keyframe $I^l_k$.
        We then compute the difference maps between the background-registered frames and the keyframe.
        The background reconstruction network takes as input the background-registered frames $\{\tilde{I}_{B, j \rightarrow k}^l\}$, 
        the difference maps $\{D^l_{B,j \rightarrow k}\}$, 
        the invalid masks $\{M^l_{B,j \rightarrow k}\}$, 
        the upsampled background $(B^{l-1}_{k})\uparrow_2$,  
        the reflection layers $(R^{l-1}_{k})\uparrow_2$, and learns to predict the residual map of the background keyframe.
        We add the predicted residual map to the upsampled background frame $(B^{l-1}_{k})\uparrow_2$ and produce the reconstructed background frame $B^{l}_{k}$ at level $l$.
        For the reflection layer reconstruction, we use the same architecture but learn a different set of network parameters.
        }
    }
    \label{fig:imageRecons}
    \figmargin
\end{figure*}

\subsection{Background/Reflection Layer Reconstruction}
\label{background_reflection_layer_reconstruction}
The layer reconstruction module aims to reconstruct the clean background image $B_k$ and the reflection image $R_k$. 
Although the two tasks of background and reflection reconstruction are similar in their goals, the characteristics of the background and reflection layers are quite different.
For example, the background layers are often more dominant in appearance but could be occluded in some frames.
On the other hand, the reflection layers are often blurry and darker.
Consequently, we train two independent networks for reconstructing the background and reflection layers.
The two networks have the same architecture but do not share the network parameters.
In the following, we only describe the network for background layer reconstruction; the reflection layer is reconstructed in a similar fashion.

We reconstruct the background layer in a coarse-to-fine fashion.
At the coarsest level ($l = 0$), we first use the flow fields estimated from the initial flow decomposition module to align the neighboring frames.
Then, we compute the average of all the background-registered frames as the predicted background image:
\begin{equation} \label{eq:init_B}
     B^0_k = \frac{1}{T} \sum_{j=1}^{T} \boldsymbol{W}(I^0_j, V^0_{B, j\rightarrow k}),
\end{equation}
where $I^0_j$ is the frame $j$ downsampled to level 0, and $\boldsymbol{W}()$ is the bilinear sampling operation.

At the $l$-th level, the network takes as input the reconstructed background image $B_k^{l-1}$, reflection image $R_k^{l-1}$, background optical flows $\{V^{l-1}_{B, k\rightarrow j}\}$ from the previous level as well as the input frames $\{I^l_t\}$ at the current level.
The model aims to reconstruct the background image of the keyframe $B_k^{l}$ at the current level.
We first upsample the background flow fields $\{V^{l-1}_{B, k\rightarrow j}\}$ by $2\times$ and align all the input frames $\{I_j^l\}$ to the keyframe $\{I_k^l\}$:
\begin{equation} \label{eq:warp}
     \tilde{I}^l_{B, j\rightarrow k} = \boldsymbol{W}( I^l_j, (V^{l-1}_{B, j\rightarrow k})\uparrow_2 ),
\end{equation}
where $()\uparrow_2$ denotes the $2\times$ bilinear upsampling operator.
As some pixels may become invalid due to occlusion or the warping from outside image boundaries, we also compute a difference map $D^l_{B,j\rightarrow k} = | I^l_{B, j\rightarrow k} - I^l_k |$ and a warping invalid masks $M^l_{B, j\rightarrow k}$ as additional cues for the network to reduce the warping artifacts.

We concatenate the registered frames, difference maps, invalid masks, and the upsampled background and reflection layers from the previous level as the input feature to the background reconstruction network.
The network then reconstructs a background image $B_k^l$ via residual learning:
\begin{align} \label{eq:background_image_reconstruction}
     B^l_k = g_B  \Big( & \{\tilde{I}_{B, j \rightarrow k}^l\}, \{D^l_{B,j \rightarrow k}\}, \{M^l_{B,j \rightarrow k}\}, (B^{l-1}_{k})\uparrow_2, \nonumber \\
     & (R^{l-1}_{k})\uparrow_2  \Big) + (B^{l-1}_{k})\uparrow_2,
\end{align}
where $g_B$ is the background reconstruction network.
Note that the reflection layer is also involved in the reconstruction of the background layer, which couples the background and reflection reconstruction networks together for joint training.
\figref{imageRecons} illustrates an overview of the background reconstruction network at the $l$-th level.
The detailed network configuration is provided in the supplementary material.

\subsection{Optical Flow Refinement}
\label{optical_flow_refinement}
After reconstructing all the background images $B^l$, we then learn to refine the background optical flows.
We use the pre-trained PWC-Net~\cite{sun2018pwc} to estimate the flow fields between a paired of background images:
\begin{equation} \label{eq:pwcnet1}
    V^l_{B, j\rightarrow k} = \text{PWC}(B^l_j, B^l_k),
\end{equation}
where $\text{PWC}$ is the pre-trained PWC-Net.
Note that the PWC-Net is fixed and not updated with the other sub-modules of our model.

\subsection{Network Training}
To improve training stability, we employ a two-stage training procedure.
At the first stage, we train the initial flow decomposition network with the following loss:
\begin{equation} \label{eq:decomposition_loss}
\begin{split}
    \mathcal{L}_{\text{dec}} = \sum_{k=1}^{T} \sum_{j=1, j\neq k}^{T} &\|V^0_{B, j\rightarrow k} - \text{PWC}(\hat{B}_j, \hat{B}_k){\downarrow}^{2^L} \|_1 + \\
    &\|V^0_{R, j\rightarrow k} - \text{PWC}(\hat{R}_j, \hat{R}_k){\downarrow}^{2^L} \|_1\,,
\end{split}
\end{equation}
where ${\downarrow}$ is the bilinear downsampling operator, $\hat{B}$ and $\hat{R}$ denote the ground-truth background and reflection layers, respectively. 
We use the pre-trained PWC-Net to compute optical flows and downsample the flows by $2^L\times$ as the ground-truth to train the initial flow decomposition network.

Next, we freeze the initial flow decomposition network and train the layer reconstruction networks with an image reconstruction loss:
\begin{equation} \label{eq:l1_loss}
    \mathcal{L}_{\text{img}} =  \frac{1}{T\!\times\!L} \sum_{t=1}^{T} \sum_{l=0}^{L}  ( \|\hat{B}^l_t - B^l_t\|_1 + \|\hat{R}^l_t - R^l_t\|_1),
\end{equation}
and a gradient loss:
\begin{equation} \label{eq:grdient_loss}
    \mathcal{L}_{\text{grad}} = \frac{1}{T\!\times\!L }\sum_{t=1}^{T} \sum_{l=0}^{L}  ( \|\nabla\hat{B}^l_t - \nabla B^l_t\|_1 + \|\nabla\hat{R}^l_t - \nabla R^l_t\|_1),
\end{equation}
where $\nabla$ is the spatial gradient operator.
The gradient loss encourages the network to reconstruct faithful edges to further improve visual quality.
The overall loss for training the layer reconstruction networks is:
\begin{equation}
    \mathcal{L} = \mathcal{L}_{\text{img}} + \lambda_{grad} \mathcal{L}_{\text{grad}},
    \label{eq:validation}
\end{equation}
where the weight $\lambda_{grad}$ is empirically set to 1 in all our experiments.
We train both the initial flow decomposition and layer reconstruction networks with the Adam optimizer~\cite{kingma2014adam} with a batch size of 2.
We set the learning rate to $10^{-4}$ for the first 100k iterations and then decrease to $10^{-5}$ for another 100k iterations.

\renewcommand{\algorithmicrequire}{\textbf{Input:}}  %
\renewcommand{\algorithmicensure}{\textbf{Output:}} %

\subsection{Synthetic Sequence Generation}
\label{sec:data_generation}
Since collecting real sequences with ground-truth reflection and background layers is very difficult, we use the Vimeo-90k dataset~\cite{xue2019video} to synthesize sequences for training.
Out of the 91,701 sequences in the Vimeo-90k training set, we randomly select two sequences as the background and reflection layers.
First, we warp the sequences using random homography transformations.
We then randomly crop the sequences to a spatial resolution of $320\times192$ pixels.
Finally, the composition is applied frame by frame using the realistic reflection image synthesis model proposed by previous work~\cite{fan2017generic, zhang2018single}.
\CR{More details about the synthetic data generation are provided in the supplementary material.}

\subsection{Online Optimization}
We observe that the model trained on our synthetic dataset may not perform well on real-world sequences.
Therefore, we propose an online refinement method to fine-tune our pre-trained model with real sequences by optimizing an unsupervised warping consistency loss:
\begin{equation} \label{eq:warp_loss}
\begin{split}
    \mathcal{L}_{\text{warp}} =   \sum_{k=1}^{T} \sum_{j=0, j \ne k}^{T} \sum_{l=0}^{L} \|I^l_j - &(\boldsymbol{W}(B^l_k, V^l_{B, j\rightarrow k}) + \\ &\boldsymbol{W}(R^l_k, V^l_{R, j\rightarrow k}) )\|_1.
\end{split}
\end{equation}
The consistency loss enhances fidelity by enforcing that the predicted background and reflection layers should be warped back and composited into the original input frames.
In addition, we also incorporate the total variation loss:
\begin{equation} \label{eq:tv_loss}
\begin{split}
    \mathcal{L}_{tv} =  \sum_{t=1}^{T} \sum_{l=0}^{L}  (\|\nabla B^l_t\|_1 + \|\nabla R^l_t\|_1),
\end{split}
\end{equation}
which encourages the network to generate natural images by following the sparse gradient image prior.
The overall loss of online optimization is:
\begin{equation} \label{eq:online_loss}
\begin{split}
    \mathcal{L}_{online} = \mathcal{L}_{\text{warp}} + \lambda_{tv} \mathcal{L}_{tv},
\end{split}
\end{equation}
where the weight $\lambda_{tv}$ is empirically set to 0.1 in all our experiments.
\CR{Note that we freeze the weight of the PWC-Net and only update the background/reflection layer reconstruction modules.}
We fine-tune our model on every single input sequence for 1k iterations, which takes about 20 minutes for a sequence with a 1296 $\times$ 864 spatial resolution. 
We use only five frames in the sequence for fine-tuning.

\subsection{Extension to Other Obstruction Removal}
The proposed framework can be easily modified to handle other obstruction removal tasks, such as fence or raindrop removal.
First, we remove the image reconstruction network for the obstruction (i.e., reflection) layer and only predict the background layers.
Second, the background image reconstruction network outputs an additional channel as the alpha map for segmenting the obstruction layer.
We do not estimate flow fields for the obstruction layer as the flow estimation network cannot handle the repetitive structures (e.g., fence) or tiny objects (e.g., raindrops) well and often predicts noisy flows.
With such a design change, our model is able to perform well on the fence and raindrop removal tasks.
We use the fence segmentation dataset~\cite{du2018accurate} and alpha matting dataset~\cite{xu2017deep} to train our model for both tasks.

\section{Experiments and Analysis}
\label{sec:experiments}

We present the main findings in this section and include more results in the supplementary material.

\begin{table*}
\caption{
        \textbf{Quantitative comparison of reflection removal methods on synthetic sequences.}
        \CR{We compare the proposed method with existing reflection removal approaches on a synthetic dataset with 100 sequences, where each sequence contains five consecutive frames.
        For the single-image based methods~\cite{fan2017generic, jin2018learning, wei2019single, yang2018seeing, zhang2018single}, we generate the results frame-by-frame.
        For multi-frame algorithms~\cite{alayrac2019visual, guo2014robust, li2013exploiting} and our method, we use five input frames to generate the results.}
    }
    \figmargin
\label{tab:compare_stoa_reflection}
\centering
\resizebox{\linewidth}{!}{
\begin{tabular}{l|l|l|cccc|cccc}
\toprule
\multicolumn{3}{l|}{\multirow{2}{*}{Method}} & \multicolumn{4}{c|}{Background} & \multicolumn{4}{c}{Reflection} \\
\multicolumn{3}{l|}{}                        & PSNR $\uparrow$ & SSIM $\uparrow$ & NCC $\uparrow$ & LMSE $\downarrow$ & PSNR $\uparrow$ & SSIM $\uparrow$ & NCC $\uparrow$ & LMSE $\downarrow$ \\
\midrule
\multirow{4}{*}{Single image} & CEILNet~\cite{fan2017generic} & CNN-based & 20.35 & 0.7429 & 0.8547 & 0.0277 & - & - & - & - \\
& Zhang~\etal~\cite{zhang2018single} & CNN-based & 19.53 & 0.7584 & 0.8526 & 0.0207 & 18.69 & 0.4945 & 0.6283 & \textcolor{blue}{\underline{0.1108}} \\
& BDN~\cite{yang2018seeing} & CNN-based & 17.08 & 0.7163 & 0.7669 & 0.0288 & - & - & - & - \\
& ERRNet~\cite{wei2019single} & CNN-based & 22.42 & \textcolor{blue}{\underline{0.8192}} & 0.8759 & \textcolor{blue}{\underline{0.0177}} & - & - & - & - \\
& Jin~\etal~\cite{jin2018learning} & CNN-based & 18.65 & 0.7597 & 0.7872 & 0.0218 & 11.44 & 0.3607 & 0.4606 & 0.1150 \\
\midrule
\multirow{4}{*}{Multiple images}            & Li and Brown~\cite{li2013exploiting} & Optimization-based & 17.12 & 0.6367 & 0.6673 & 0.0604 & 7.68 & 0.2670 & 0.3490 & 0.1214     \\
& Guo~\etal~\cite{guo2014robust} & Optimization-based & 14.58 & 0.5077 & 0.5802 & 0.0694 & 14.12 & 0.3150 & 0.3516 & 0.1774 \\
& Alayrac~\etal~\cite{alayrac2019visual} & CNN-based & \textcolor{blue}{\underline{23.62}} & 0.7867 & \textcolor{blue}{\underline{0.9023}} & 0.0200 & \textcolor{blue}{\underline{21.18}} & \textcolor{blue}{\underline{0.6320}} & \textcolor{blue}{\underline{0.7535}} & 0.1517 \\
& Ours w/o online optim. & CNN-based & \textcolor{red}{\pmb{26.57}} & \textcolor{red}{\pmb{0.8676}} & \textcolor{red}{\pmb{0.9380}} & \textcolor{red}{\pmb{0.0125}} & \textcolor{red}{\pmb{21.42}} & \textcolor{red}{\pmb{0.6438}} & \textcolor{red}{\pmb{0.7613}} & \textcolor{red}{\pmb{0.1008}} \\
\bottomrule
\end{tabular}
}
\end{table*}

\begin{figure}
\centering
\footnotesize
\renewcommand{\tabcolsep}{1pt} %
\renewcommand{\arraystretch}{1} %
\begin{tabular}{ccc}
    \makecell{Input\\(rep. frame)} &\makecell{ Recovered\\background} & \makecell{Recovered\\obstruction} \\
    \includegraphics[width=0.31\columnwidth]{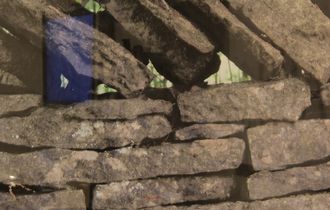} & \includegraphics[width=0.31\columnwidth]{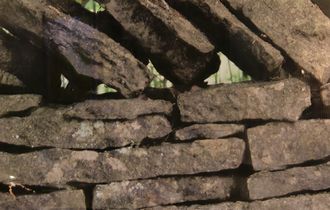} & \includegraphics[width=0.31\columnwidth]{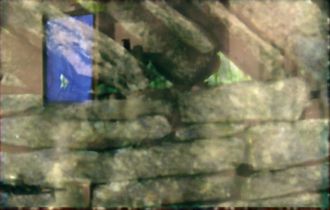} \\
    Stone & NCC = 0.9660 & NCC = 0.7006 \\
    \includegraphics[width=0.31\columnwidth]{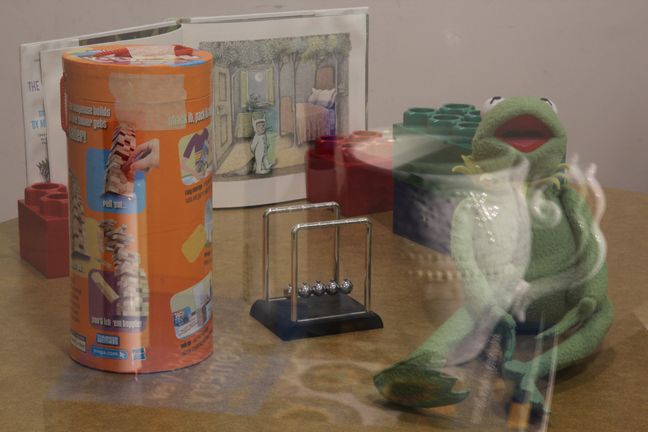} & \includegraphics[width=0.31\columnwidth]{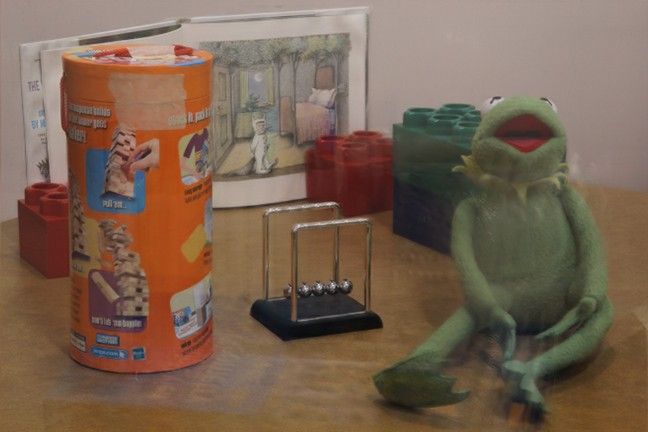} & \includegraphics[width=0.31\columnwidth]{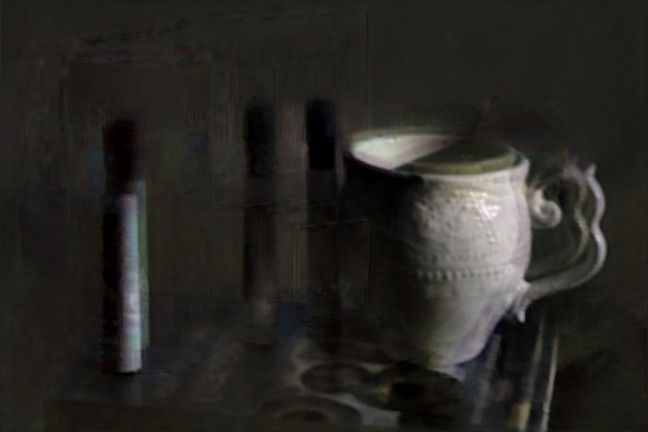} \\
    Toy & NCC = 0.9487 & NCC = 0.8707 \\ 
    \includegraphics[width=0.31\columnwidth]{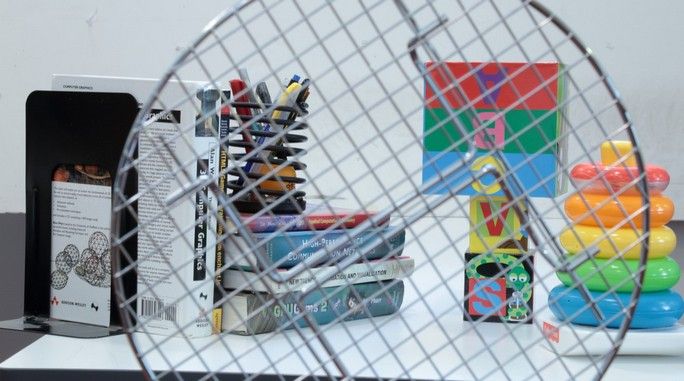} & \includegraphics[width=0.31\columnwidth]{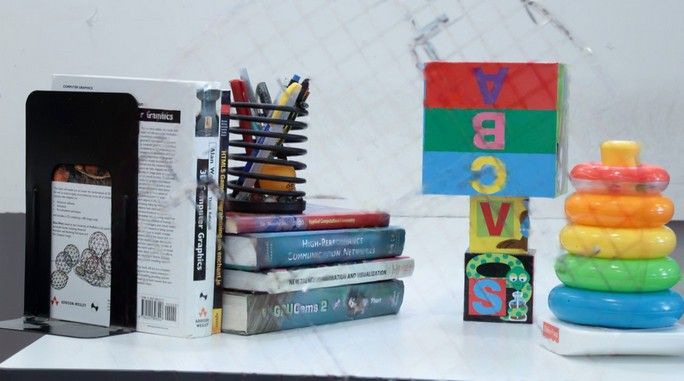} & \includegraphics[width=0.31\columnwidth]{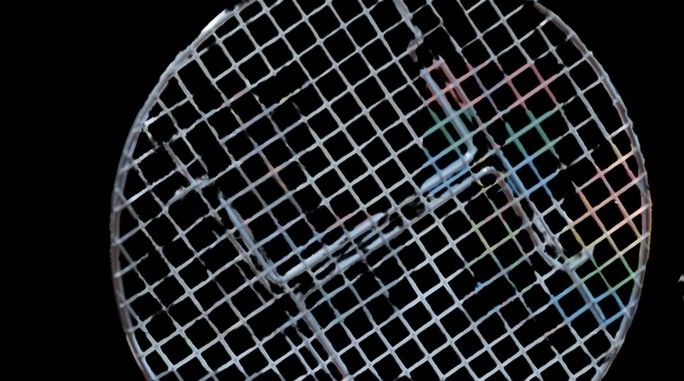} \\
    Hanoi & NCC = 0.9938 & NCC = 0.8267 \\ 
\end{tabular}
\vspace{1mm}
\renewcommand{\tabcolsep}{3pt} %
\renewcommand{\arraystretch}{1} %
\begin{tabular}{lcccccc}
\toprule
\multirow{2}{*}{Method} & \multicolumn{2}{c}{Stone} & \multicolumn{2}{c}{Toy} & \multicolumn{2}{c}{Hanoi} \\
& $B$ & $R$ & $B$ & $R$ & $B$ & $O$ \\
\midrule
Li and Brown~\cite{li2013exploiting}& 0.9271 & 0.2423 &0.7906 & 0.6084 & - & - \\
Guo~\etal~\cite{guo2014robust}&0.7258& 0.1018& 0.7701& 0.6860 & - & -\\
Xue~\etal~\cite{xue2015computational}&\textcolor{red}{\pmb{0.9738}}& \textcolor{red}{\pmb{0.8433}}& \textcolor{blue}{\underline{0.8985}}& \textcolor{blue}{\underline{0.7536}} & \textcolor{blue}{\underline{0.9921}} & \textcolor{blue}{\underline{0.7079}} \\
Alayrac~\etal~\cite{alayrac2019visual}& 0.9367 & 0.1633 & 0.7985 & 0.5263  & - & -\\
Ours & \textcolor{blue}{\underline{0.9660}} & \textcolor{blue}{\underline{0.7006}} & \textcolor{red}{\pmb{0.9487}} & \textcolor{red}{\pmb{0.8707}}  & \textcolor{red}{\pmb{0.9938}} & \textcolor{red}{\pmb{0.8267}}\\
\bottomrule
\end{tabular}
\figmargin
\caption{\textbf{Quantitative evaluation on controlled sequences.} For each sequence, we show the keyframe (\emph{left}) and recovered background (\emph{middle}) and reflection/occluder (\emph{right}). We report the NCC scores of recovered backgrounds and reflections for quantitative comparisons.
}
\label{fig:controlled}
\end{figure}

\subsection{Comparisons with State-of-the-arts}

\heading{Controlled sequences.}
We first evaluate on the controlled sequences provided by Xue et al~\cite{xue2015computational}, which contain three videos with ground-truth background and reflection layers.
We compare the proposed method with Li and Brown~\cite{li2013exploiting}, Guo et al.~\cite{guo2014robust}, Xue et al.~\cite{xue2015computational}, and Alayrac et al.~\cite{alayrac2019visual}.
\figref{controlled} shows our recovered background and reflection/obstruction layers and the normalized cross-correlation (NCC) scores~\cite{wan2017benchmarking, xue2015computational}.
Our method performs favorably against other approaches on the Toy and Hanoi sequences and shows comparable scores to Xue et al.~\cite{xue2015computational} on the Stone sequence.

\heading{Synthetic sequences.}
We synthesize 100 sequences by the method described in~\secref{data_generation} from the Vimeo-90k test set.
We compare our approach with five single-image reflection removal methods~\cite{fan2017generic,jin2018learning,wei2019single,yang2018seeing,zhang2018single}, and three multi-frame approaches~\cite{alayrac2019visual, guo2014robust,li2013exploiting}.
We use the default parameters of each method to generate the results.
Since Alayrac et al.~\cite{alayrac2019visual} do not release the source code or pre-trained model, we re-implement their model and train on our training dataset.
\tabref{compare_stoa_reflection} shows the average PSNR, SSIM~\cite{wang2004image}, NCC, and LMSE~\cite{grosse2009ground} metrics.
The proposed method obtains the best scores on all the evaluation metrics for both background and reflection layers.

\begin{figure*}
\centering
\footnotesize
\renewcommand{\tabcolsep}{1pt} %
\renewcommand{\arraystretch}{1} %
\newcommand{\imagewidth}{0.325\columnwidth}
\begin{tabular}{ccccccc}
    \includegraphics[width=\imagewidth]{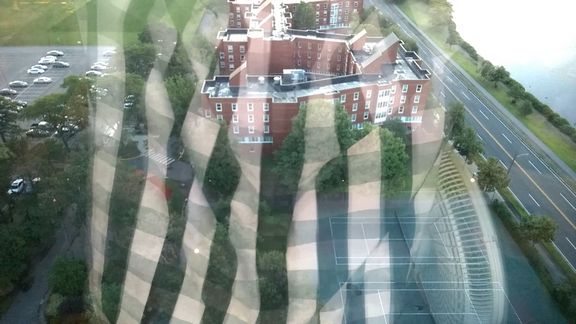} & 
    \raisebox{2.0\normalbaselineskip}[0pt][0pt]{\rotatebox[origin=c]{90}{Background}} &  
    \includegraphics[width=\imagewidth]{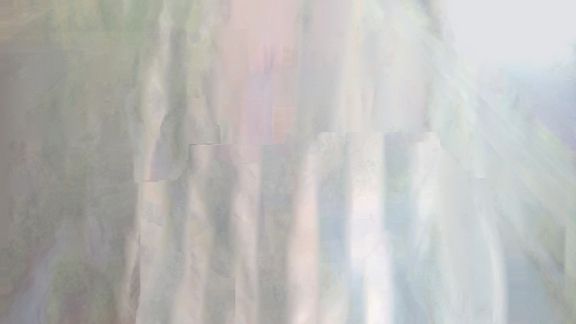} &  
    \includegraphics[width=\imagewidth]{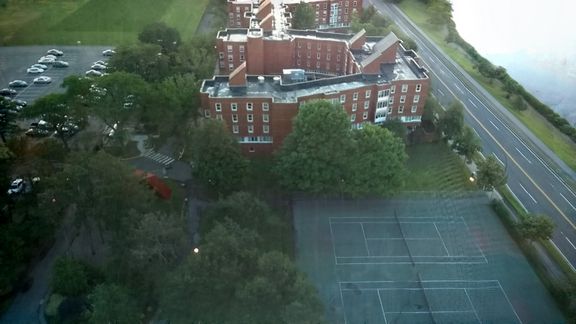} &  
    \includegraphics[width=\imagewidth]{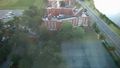} &  
    \includegraphics[width=\imagewidth]{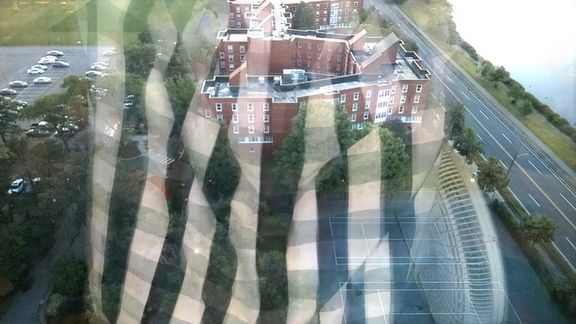} & 
    \includegraphics[width=\imagewidth]{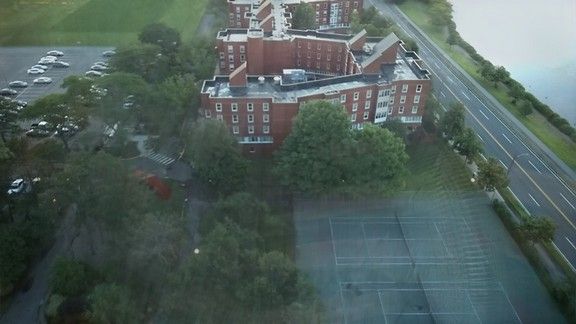} 
    \\
    & 
    \raisebox{2.0\normalbaselineskip}[0pt][0pt]{\rotatebox[origin=c]{90}{Reflection}} &  
    \includegraphics[width=\imagewidth]{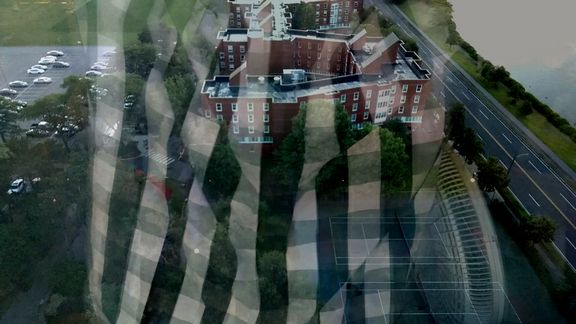} &  
    \includegraphics[width=\imagewidth]{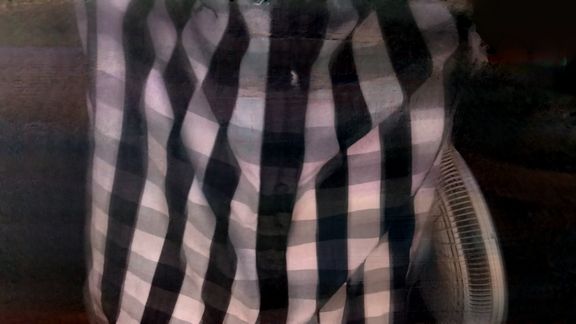} &  
    \includegraphics[width=\imagewidth]{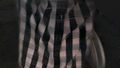} &  
    \includegraphics[width=\imagewidth]{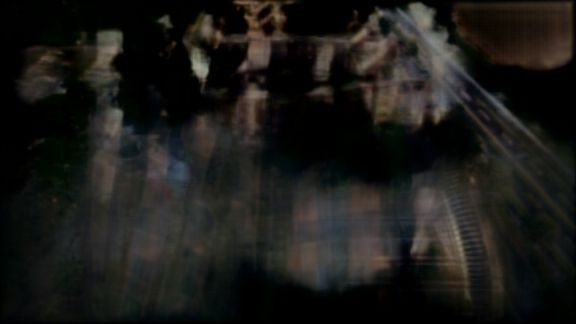} & 
    \includegraphics[width=\imagewidth]{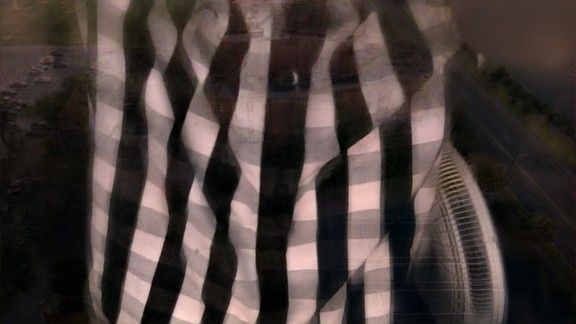} 
    \\
    \includegraphics[width=\imagewidth]{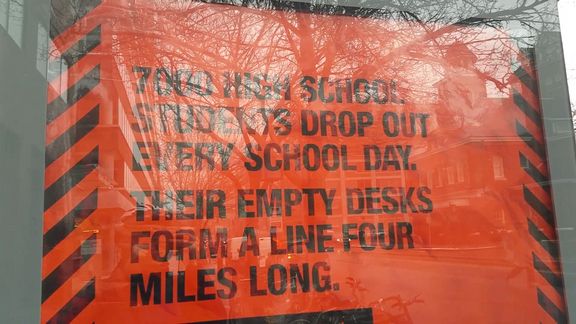} & 
    \raisebox{2.0\normalbaselineskip}[0pt][0pt]{\rotatebox[origin=c]{90}{Background}} &  
    \includegraphics[width=\imagewidth]{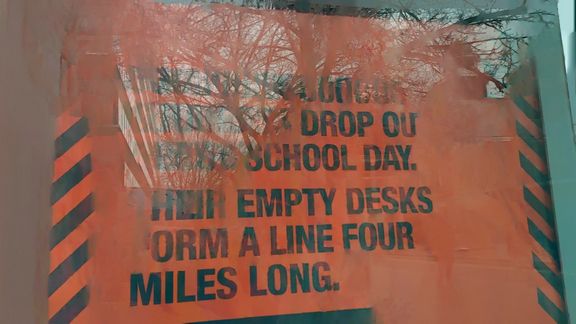} &  
    \includegraphics[width=\imagewidth]{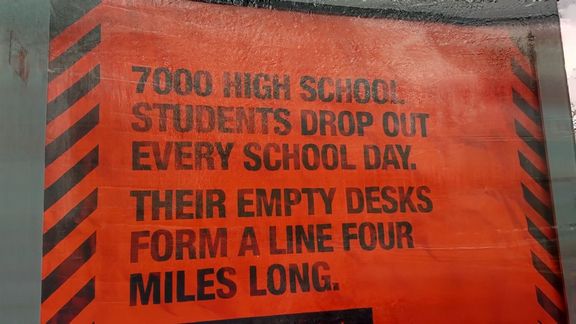} &  
    \includegraphics[width=\imagewidth]{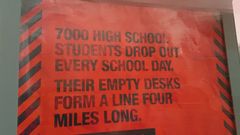} &  
    \includegraphics[width=\imagewidth]{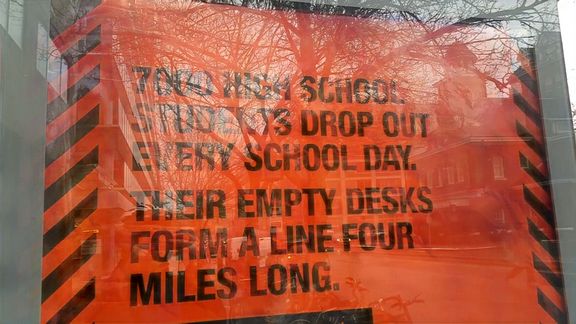} & 
    \includegraphics[width=\imagewidth]{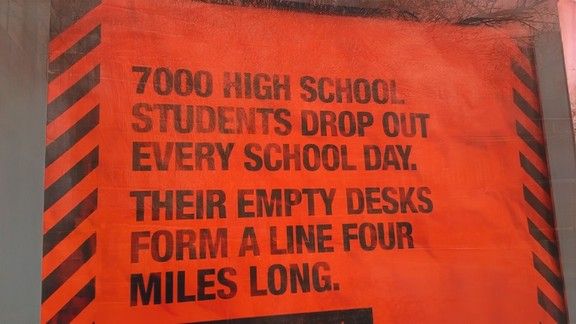} 
    \\
    & 
    \raisebox{2.0\normalbaselineskip}[0pt][0pt]{\rotatebox[origin=c]{90}{Reflection}} &  
    \includegraphics[width=\imagewidth]{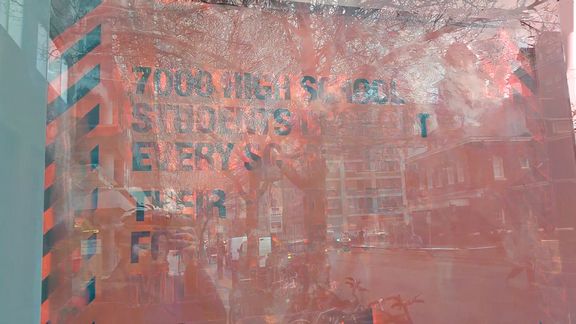} &  
    \includegraphics[width=\imagewidth]{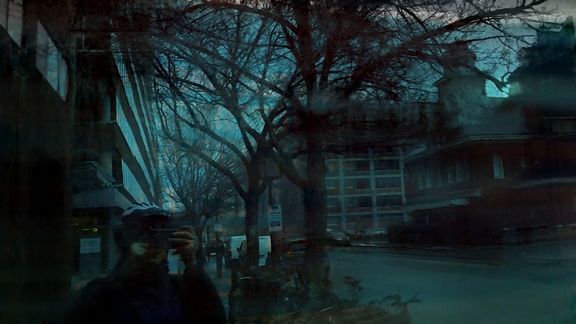} &  
    \includegraphics[width=\imagewidth]{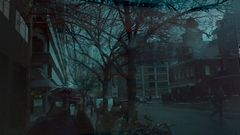} &  
    \includegraphics[width=\imagewidth]{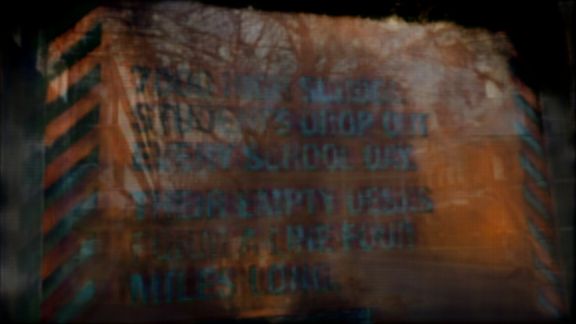} & 
    \includegraphics[width=\imagewidth]{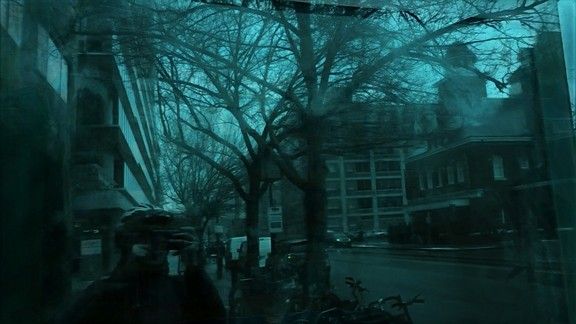} 
    \\
    Representative input frame & & Li and Brown~\cite{li2013exploiting} & Xue~\etal~\cite{xue2015computational} & Nandoriya~\etal~\cite{nandoriya2017video}* & Alayrac~\etal~\cite{alayrac2019visual} & Ours 
\end{tabular}
\figmargin
\caption{\textbf{Visual comparison of background-reflection separation on natural sequences.} More results can be found in the supplementary material. *Results are in lower resolution.
} 
\label{fig:reflection_visual_1}
\end{figure*}

\begin{figure}
\centering
\footnotesize
\renewcommand{\tabcolsep}{1pt} %
\renewcommand{\arraystretch}{1} %
\newcommand{\imagewidth}{0.32\columnwidth}
\newcommand{\patchwidth}{0.155\columnwidth}
\begin{tabular}{cccccc}
    \multicolumn{2}{c}{\includegraphics[width=\imagewidth]{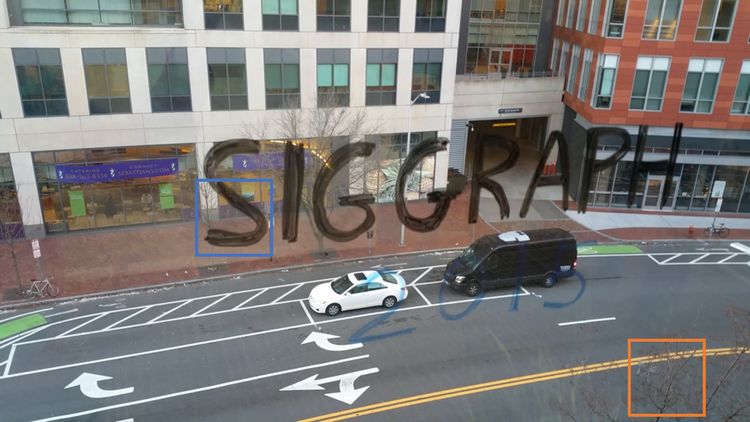}} & 
    \multicolumn{2}{c}{\includegraphics[width=\imagewidth]{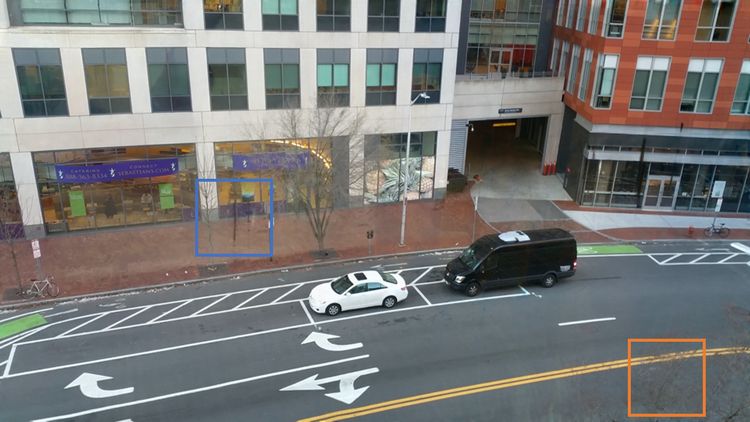}} &  
    \multicolumn{2}{c}{\includegraphics[width=\imagewidth]{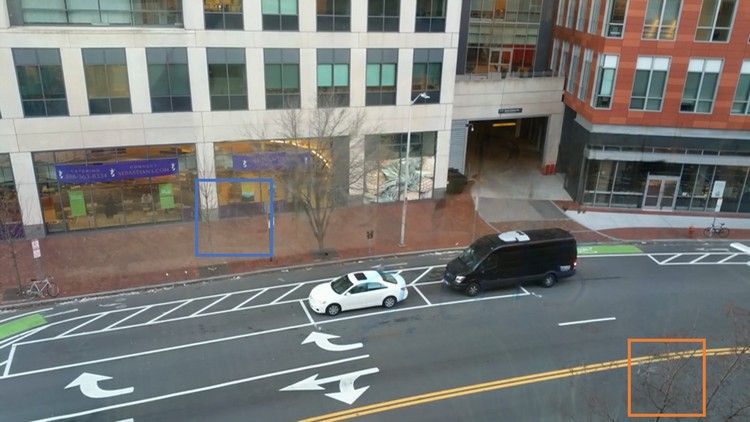}} 
    \\
    \includegraphics[width=\patchwidth]{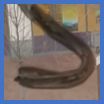} & 
    \includegraphics[width=\patchwidth]{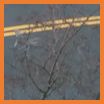} &
    \includegraphics[width=\patchwidth]{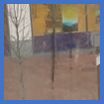} & 
    \includegraphics[width=\patchwidth]{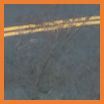} & 
    \includegraphics[width=\patchwidth]{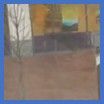} & 
    \includegraphics[width=\patchwidth]{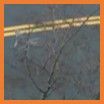} 
    \\
    \multicolumn{2}{c}{\includegraphics[width=\imagewidth]{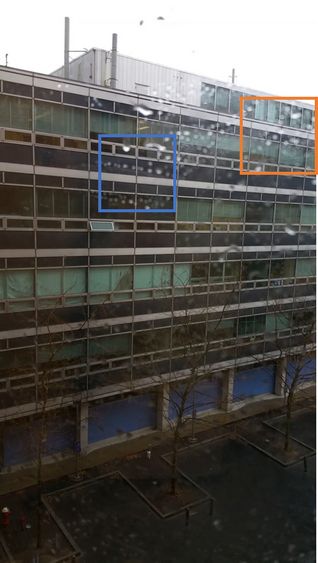}} &
    \multicolumn{2}{c}{\includegraphics[width=\imagewidth]{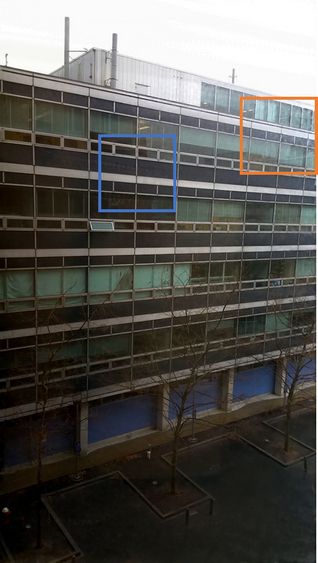}} &
    \multicolumn{2}{c}{\includegraphics[width=\imagewidth]{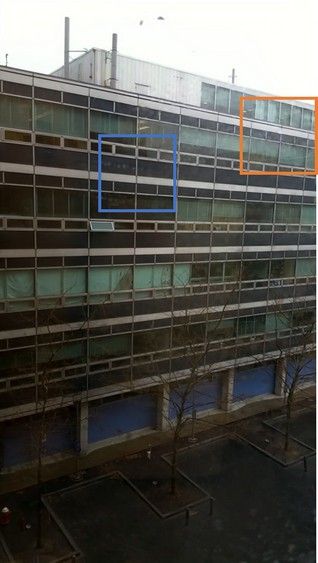}} 
    \\
    \includegraphics[width=\patchwidth]{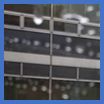} & 
    \includegraphics[width=\patchwidth]{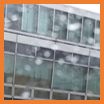} &
    \includegraphics[width=\patchwidth]{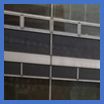} & 
    \includegraphics[width=\patchwidth]{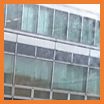} & 
    \includegraphics[width=\patchwidth]{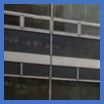} & 
    \includegraphics[width=\patchwidth]{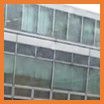} 
    \\
    \multicolumn{2}{c}{\makecell{Representative input \\frame}} &
    \multicolumn{2}{c}{Xue~\etal~\cite{xue2015computational}} & 
    \multicolumn{2}{c}{Ours} \\
\end{tabular}
\figmargin
\caption{\textbf{Recovering occluded scenes by raindops.} }
\label{fig:obstruction_visual_2}
\figmargin
\end{figure}

\heading{Real sequences.}
In~\figref{reflection_visual_1}, we present visual comparisons of real input sequences from~\cite{xue2015computational}.
Our method is able to separate the reflection layers and reconstruct clear and sharp background images than other approaches~\cite{alayrac2019visual, li2013exploiting, nandoriya2017video, xue2015computational}.
\figref{obstruction_visual_2} shows two examples where the inputs contain obstruction such as texts on the glass or raindrops.
Our method can remove the obstruction layer and reconstruct clear background images.
More visual comparisons are available in the supplementary material.

\subsection{Analysis and Discussion}

In this section, we analyze several key design choices of the proposed framework.
We also provide the execution time and show a failure case of our method.

\heading{Initial flow decomposition.}
We demonstrate that the uniform flow initialization plays an important role in our algorithm.
We train our model with the following settings: 1) removing the initial flow decomposition network, where the flows at the coarsest level are set to zero, and 2) predicting spatially-varying dense flow fields as the initial flows.
\tabref{ablation}(a) reports the validation loss of~\eqnref{validation} on our Vimeo-90k validation set, where the model with uniform flow prediction achieves a much lower validation loss compared to the alternatives.
Initializing the flow fields to zero makes it difficult for the following levels to decompose the background and reflection layers.
On the contrary, estimating dense flow fields at the coarsest level may result in noisy predictions and lead to inconsistent layer separation.
Our uniform flow prediction strikes a balance and serves as a good initial prediction to facilitate the following background reconstruction and flow refinement steps.

\heading{Image reconstruction network.}
To demonstrate the effectiveness of the image reconstruction network, we replace it with a temporal filter to fuse the neighbor frames, which are warped and aligned by the optical flows.
We show in~\tabref{ablation}(b) that both the temporal mean and median filters result in large errors (in terms of the validation loss of~\eqnref{validation}) as the errors are accumulated across levels.
In contrast, our image reconstruction network learns to reduce warping and alignment errors and generates clean foreground and background images.

\begin{figure*}
\centering
\footnotesize
\renewcommand{\tabcolsep}{3pt} %
\renewcommand{\arraystretch}{1} %
\newcommand{\imagewidth}{0.19\linewidth}
\begin{tabular}{ccccc}
    \includegraphics[width=\imagewidth]{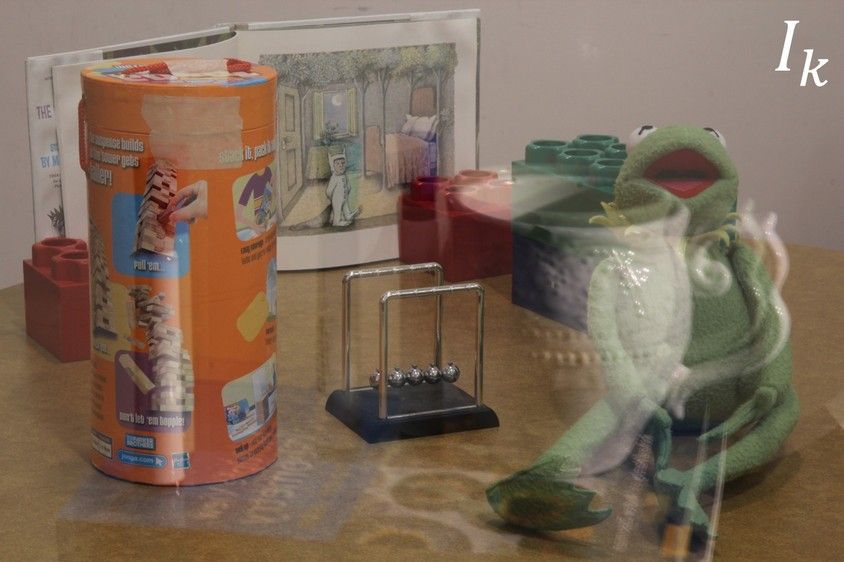} & 
    \raisebox{2.6\normalbaselineskip}[0pt][0pt]{\rotatebox[origin=c]{90}{Background}} &
    \includegraphics[width=\imagewidth]{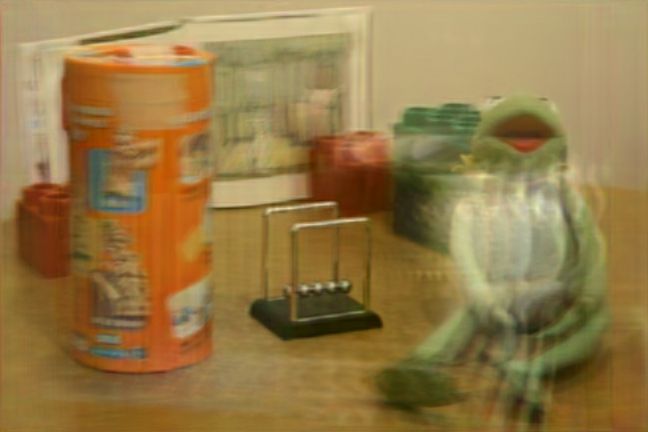} & 
    \includegraphics[width=\imagewidth]{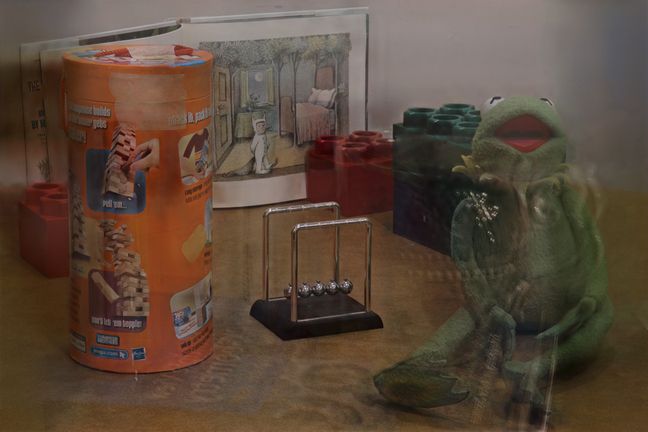} &  
    \includegraphics[width=\imagewidth]{Figures/results/reflection/Ours_online/00001B2_norm.jpg} 
    \\
    & 
    \raisebox{2.6\normalbaselineskip}[0pt][0pt]{\rotatebox[origin=c]{90}{Reflection}} &  
    \includegraphics[width=\imagewidth]{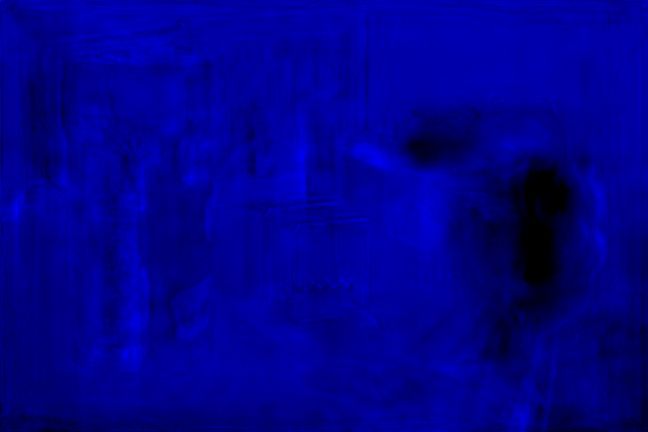} & 
    \includegraphics[width=\imagewidth]{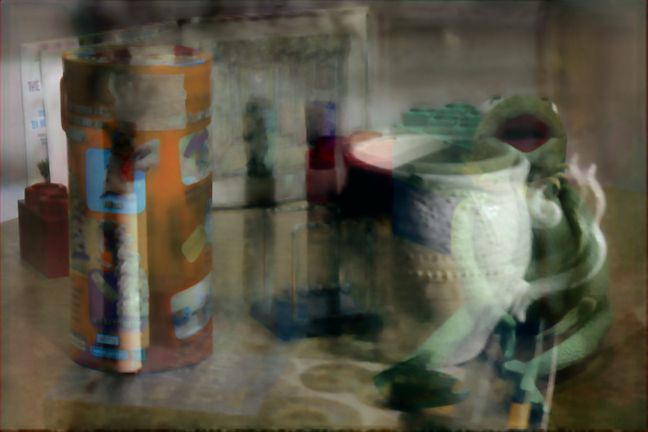} &  
    \includegraphics[width=\imagewidth]{Figures/results/reflection/Ours_online/00001F2_norm.jpg} 
    \\
    \makecell{Representative input frame} & & 
    \makecell{w/ online optimization \\ w/o pre-training} & 
    \makecell{w/o online optimization \\ w/ pre-training} & 
    \makecell{w/ online optimization \\ w/ pre-training} \\
\end{tabular}
\figmargin
\caption{\textbf{Effect of online optimization and pre-training.} Both steps are crucial to achieving high-quality results.} 
\label{fig:online}
\figmargin
\end{figure*}

\heading{Online optimization.}
\tabref{ablation}(c) shows that both the network pre-training with synthetic data and online optimization with real data are beneficial to the performance of our model.
In~\figref{online}, we show that the model without pre-training cannot separate the reflection well on the real input sequence.
Without online optimization, the background image contains residuals from the reflection layer.
After online optimization, our method is able to reconstruct both background and reflection layers well.

\begin{table*}
\caption{
        \textbf{Ablations.} We analyze the design choices of the proposed method and report the validation loss of~\eqnref{validation} on the synthetic reflection-background Vimeo-90k test set. %
    }
\label{tab:ablation}
\figmargin
\centering
\footnotesize
\begin{tabular}{p{4.5cm}p{5.7cm}p{5.5cm}}
(a) \tb{Initial flow decomposition}: Predicting uniform flow fields as initialization achieves better results. & 
(b) \tb{Fusion method}: Our image reconstruction network recovers better background/reflection than temporal mean/median filtering. & 
(c) \tb{Model training}: Both the network pre-training and online optimization are important to the performance of our method.
\end{tabular}

\figcapmargin

\begin{tabular}{p{4.5cm}p{5.7cm}p{5.5cm}}
\makecell{
    \begin{tabular}{l|c}
    \toprule
    Flow initialization & Loss \\
    \midrule
    \emph{Zero} initialization & 0.377 \\
    \emph{Dense} flow field & 0.226 \\
    \emph{Uniform} flow field (Ours) & \pmb{0.184} \\
    \bottomrule
    \end{tabular}
}
& 
\makecell{
    \begin{tabular}{l|c}
    \toprule
    Image fusion method & Loss \\
    \midrule
    Temporal mean filtering & 0.526 \\
    Temporal median filtering & 0.482 \\
    Image reconstruction network (Ours) & \pmb{0.184} \\
    \bottomrule
    \end{tabular}
} 
& 
\makecell{
    \begin{tabular}{cc|c}
    \toprule
    \makecell{Online optimization} & Pre-training & \makecell{Loss} \\
    \midrule
    \checkmark & - & 0.417 \\
    - & \checkmark & 0.184 \\
    \checkmark & \checkmark & \pmb{0.139} \\
    \bottomrule
    \end{tabular}
}
\end{tabular}
\end{table*}

\heading{Running time.}
We evaluate the execution time of two optimization-based algorithms~\cite{guo2014robust, li2013exploiting} and a recent CNN-based method~\cite{alayrac2019visual} with different input sequences resolutions on a computer with Intel Core i7-8550U CPU and NVIDIA TITAN Xp GPU.
\tabref{running_time} shows that our method without the online optimization step runs faster than optimization-based algorithms.
Alayrac et al.~\cite{alayrac2019visual} use a 3D CNN architecture without explicit motion estimation, which results in a faster inference speed.
In contrast, our method computes bi-directional optical flows for every pair of input frames in a coarse-to-fine manner, which is slower but achieves much better reconstruction performance.

\begin{table}[t]
\caption{\textbf{Running time comparison (in seconds).} CPU: Intel Core i7-8550U, GPU: NVIDIA TITAN Xp. * denotes methods using GPU.}
\label{tab:running_time}
\figmargin
\centering
\footnotesize
\resizebox{\linewidth}{!}{
\begin{tabular}{lccc}
\toprule
 & \makecell{QVGA\\($320\times240$)} & \makecell{VGA\\($640\times480$)} & \makecell{720p\\($1280\times720$)} \\
\midrule
Li and Brown~\cite{li2013exploiting} & 82.591 & 388.235 & 1304.231 \\
Guo~\etal~\cite{guo2014robust} & 64.251 & 369.200 & 1129.125 \\
*Alayrac~\etal~\cite{alayrac2019visual}  & 0.549 & 2.011 & 6.327 \\
*Ours w/o online optim. & 1.107 & 2.216 & 9.857\\
*Ours w/ online optim. & 66.056 & 264.227 & 929.182\\
\bottomrule
\end{tabular}
}
\figmargin
\end{table}

\begin{figure}[t]
\centering
\footnotesize
\renewcommand{\tabcolsep}{1pt} %
\renewcommand{\arraystretch}{1} %
\newcommand{\imagewidth}{0.42\columwidth}
\begin{tabular}{ccc}
    \includegraphics[width=0.42\columnwidth]{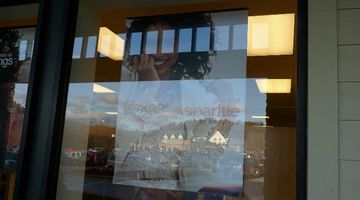} & 
    \raisebox{2.5\normalbaselineskip}[0pt][0pt]{\rotatebox[origin=c]{90}{Background}} &  
    \includegraphics[width=0.42\columnwidth]{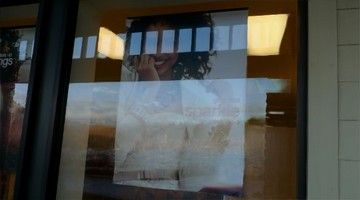} 
    \\
    \includegraphics[width=0.42\columnwidth]{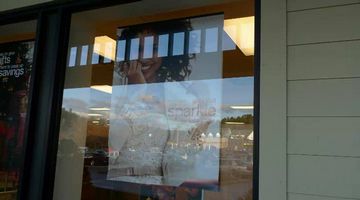} & 
    \raisebox{2.5\normalbaselineskip}[0pt][0pt]{\rotatebox[origin=c]{90}{Reflection}} &  
    \includegraphics[width=0.42\columnwidth]{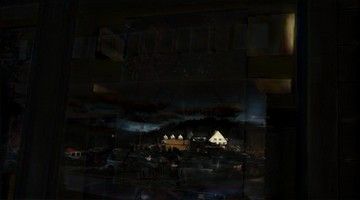} 
    \\
    Representative input frame & & Our results \\
\end{tabular}
\figmargin
\caption{\textbf{A failure case.} Our method fails to recover the correct flow fields for each layer, leading to ineffective reflection removal.}
\label{fig:failure}
\figmargin
\end{figure}

\heading{Failure case.}
We show a failure case of our algorithm in~\figref{failure}, where our method does not separate the reflection layer well.
This example is particularly challenging as there are two layers of reflections: the top part contains the wooden beams, and the bottom part comes from the street behind the camera.
As the motion of the wooden beams is close to the background image, our method can only separate the street scenes in the reflection layer.

\section{Conclusions}
\label{sec:conclusion}
We have presented a novel method for multi-frame reflections and obstructions removal.
Our key insight is to leverage a CNN to reconstruct background and reflection layers from flow-warped images.
Integrating optical flow estimation and coarse-to-fine refinement enable our model to robustly recover the underlying clean image from challenging real-world sequences.
Our method can be applied to different tasks such as fence or raindrop removal with minimum changes in our design.
We also show that online optimization on testing sequences leads to improved visual quality.
Extensive visual comparisons and quantitative evaluation demonstrate that our approach performs well on a wide variety of scenes.

\vspace{1mm}
{
\heading{Acknowledgments.} This work is supported in part by NSF CAREER ($\#$1149783), NSF CRII ($\#$1755785), MOST 109-2634-F-002-032, MediaTek Inc. and gifts from Adobe, Toyota, Panasonic, Samsung, NEC, Verisk, and Nvidia.
}

{\small
\bibliographystyle{ieee_fullname}
\bibliography{egbib}
}

\end{document}